\documentclass[letterpaper]{article} 
\usepackage{formation}  
\usepackage{times}  
\usepackage{helvet}  
\usepackage{courier}  
\usepackage[hyphens]{url}  
\usepackage{graphicx} 
\usepackage{natbib}  
\usepackage{caption} 
\usepackage{algorithm}
\usepackage{algorithmic}
\usepackage{amsmath}
\usepackage{xspace}
\usepackage{booktabs}
\usepackage{subcaption}
\usepackage{pifont}
\usepackage{amssymb}
\usepackage{color}
\def\ie{\textit{i.e.}\xspace}

\def\eg{\textit{e.g.}\xspace}

\usepackage{newfloat}
\usepackage{listings}
\DeclareCaptionStyle{ruled}{labelfont=normalfont,labelsep=colon,strut=off} 
\floatstyle{ruled}
\newfloat{listing}{tb}{lst}{}
\floatname{listing}{Listing}
\title{Optimizing Federated Graph Learning with Inherent Structural\\ Knowledge and Dual-Densely Connected GNNs}
\author{
    Longwen Wang\textsuperscript{\rm 1}, 
    Jianchun Liu\textsuperscript{\rm 2}\thanks{Corresponding author.}, 
    Zhi Liu\textsuperscript{\rm 3},
    Jinyang Huang\textsuperscript{\rm 4}
}
\affiliations {
    \textsuperscript{\rm 1}School of Computer Science and Technology, Xidian University, China\\
    \textsuperscript{\rm 2}the School of Computer Science and Technology, University of Science and Technology of China\\
    \textsuperscript{\rm 3}The University of Electro-Communications, Tokyo, Japan\\
    \textsuperscript{\rm 4}School of Computer and Information, Hefei University of Technology,China\\
    abeiduoicon@gmail.com, jcliu17@ustc.edu.cn, liu@ieee.org, hjy@hfut.edu.cn
}

\usepackage{bibentry}

\begin{document}

\maketitle

\begin{abstract}
Federated Graph Learning (FGL) is an emerging technology that enables clients to collaboratively train powerful Graph Neural Networks (GNNs) in a distributed manner without exposing their private data. 
Nevertheless, FGL still faces the challenge of
the severe non-Independent and Identically Distributed (non-IID) nature of graphs, which possess diverse node and edge structures, especially across varied domains. 
Thus, exploring the knowledge inherent in these structures becomes significantly crucial.
Existing methods, however, either overlook the inherent structural knowledge in graph data or capture it at the cost of significantly increased resource demands (\eg, FLOPs and communication bandwidth), which can be detrimental to distributed paradigms.
Inspired by this, we propose FedDense, a novel FGL framework that optimizes the utilization efficiency of inherent structural knowledge.
To better acquire knowledge of diverse and underexploited structures, FedDense first explicitly encodes the structural knowledge inherent within graph data itself alongside node features. 
Besides, FedDense introduces a Dual-Densely Connected (DDC) GNN architecture that exploits the multi-scale (\ie, one-hop to multi-hop) feature and structure insights embedded in the aggregated feature maps at each layer.
In addition to the exploitation of inherent structures, we consider resource limitations in FGL, devising exceedingly narrow layers atop the DDC architecture and adopting a selective parameter sharing strategy to reduce resource costs substantially.
We conduct extensive experiments using 15 datasets across 4 different domains, demonstrating that FedDense consistently surpasses baselines by a large margin in training performance, while demanding minimal resources.
\end{abstract}

\section{Introduction}
The rising interest in Graph Neural Networks (GNNs) is fueled by the extensive availability of graph data across various domains, such as chemical molecules \cite{qian2021directed-molecular}, bioinformatics \cite{muzio2021biological}, social networks \cite{monti2019fakesocial}, and computer vision \cite{pradhyumna2021graph-computervision}. 
Traditional GNNs require graph data to be centralized for processing and analysis. 
However, escalating privacy concerns and the increased need for cross-domain collaboration have made addressing privacy breaches and data silos crucial.
To this end, Federated Graph Learning (FGL) \cite{zhang2021FGL-survy}, which integrates Federated Learning (FL) \cite{aledhari2020federated-survey} into the training process of GNNs, has been proposed. 
FGL effectively addresses both privacy breaches and data silos through a distributed paradigm, allowing clients to collaboratively train GNNs without disclosing their private data, thereby unlocking the full potential of GNNs.

\begin{figure}[t]\centering
    \begin{subfigure}{0.48\linewidth}
                \centering
                \includegraphics[width=1\linewidth]{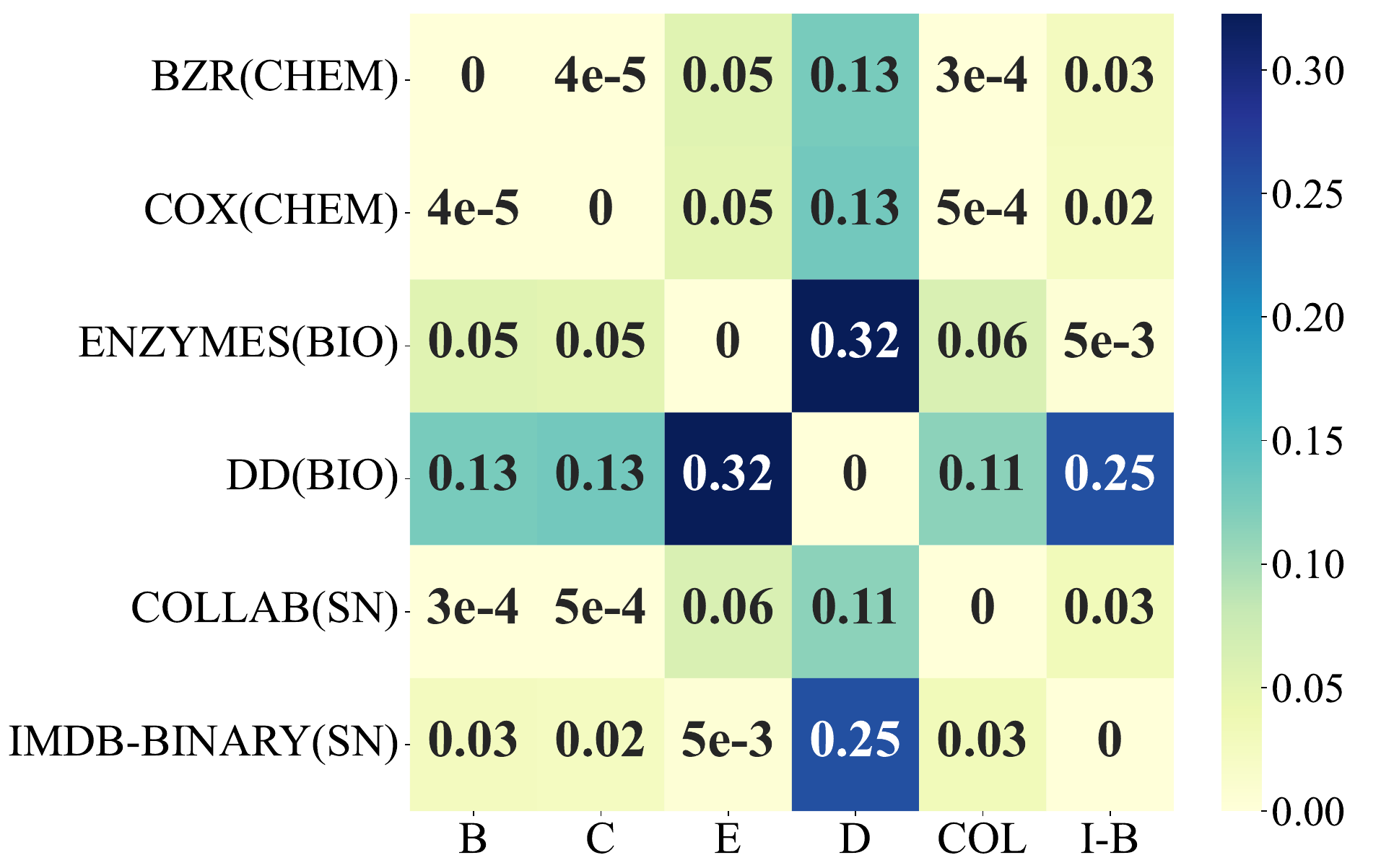}
    \end{subfigure}
    \begin{subfigure}{0.48\linewidth}
                \centering
                \includegraphics[width=1\linewidth]{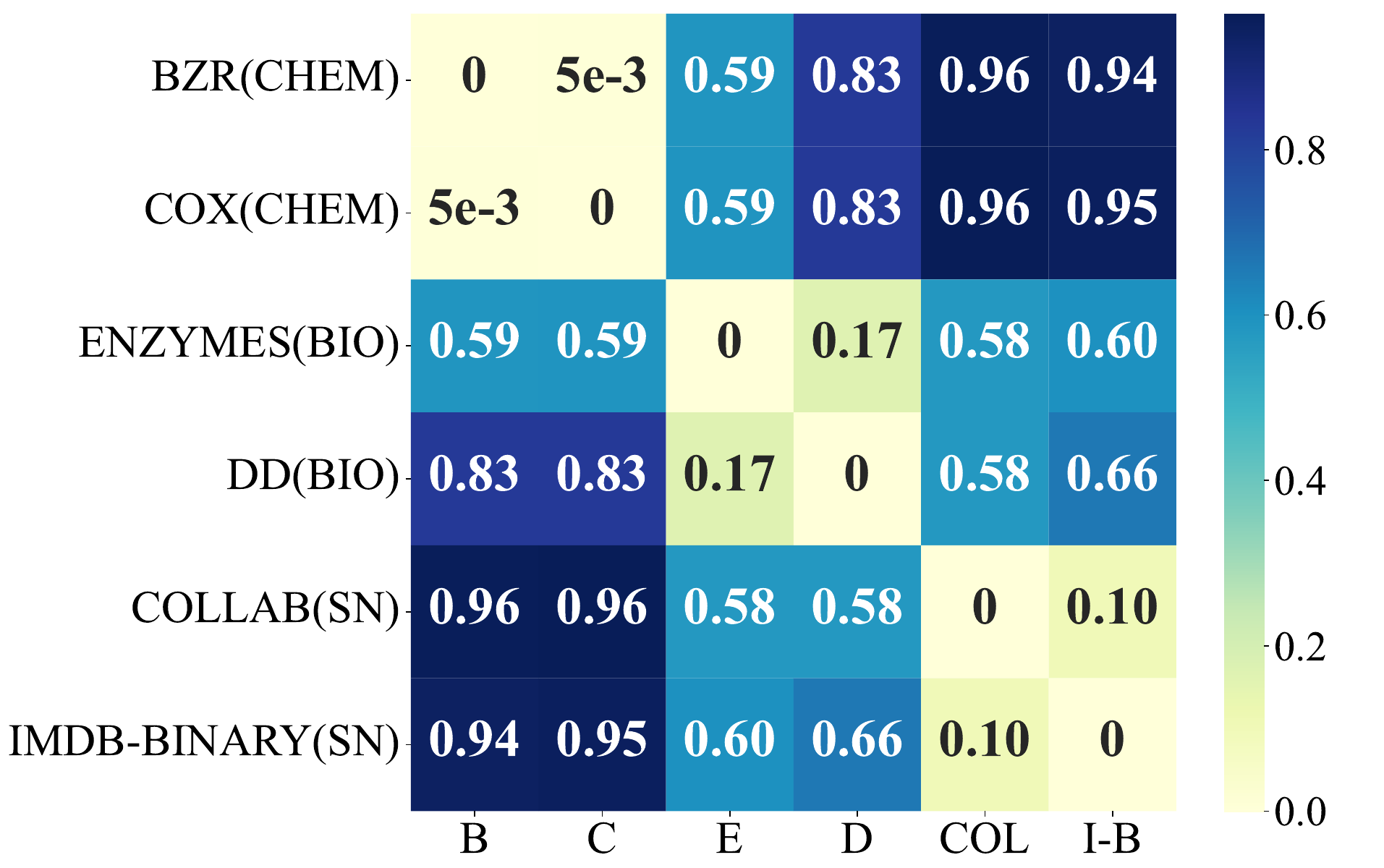}
    \end{subfigure}
    \caption{The Jensen-Shannon divergence heatmap compares feature (\emph{left}) and structure (\emph{right}) information among six graph datasets from varied domains. Feature information: empirical feature similarity distribution between all linked nodes; Structure information: concatenation of degree divergence and clustering coefficient distributions.}\label{heatmap}
\end{figure}

Existing FGL methods predominantly rely on traditional GNNs that employ a feature-based message-passing mechanism, where each graph node representation is iteratively updated by aggregating features from its one-hop to multi-hop neighbors. 
However, this feature-based mechanism overlooks the unique structural information inherent in graph data.
The structures of nodes and edges are not merely supplementary but are fundamental characteristics of graphs, embodying knowledge distinct from features \cite{buffelli2022impact-structureofgraph2,li2017locally-structureofgraph}.
As shown in Figure \ref{heatmap}, feature and structure heterogeneity across different domains exhibit completely disparate patterns. For feature information, even data from the same domain (\eg, DD (BIO) and ENZYMES (BIO)) reveal significant variation, while their structure remains quite similar. Conversely, data from different domains (\eg, Bioinformatics and Social Networks) exhibit very small feature heterogeneity but substantial differences in structural information. These findings indicate that structural knowledge plays a unique role alongside features in graph data, particularly across domains. Thus, a local model on each client that relies solely on features may fail to capture structural insights, potentially resulting in misaligned updates across clients and hindering overall model performance \cite{positionICLR2022,tan2023fedstar}.

\begin{table}[t]
\centering
{\fontsize{9}{\baselineskip}\selectfont
\begin{tabular}{@{}cccc@{}}
\toprule
Metrics             & FLOPs(G)                           & Model Size(KB)                      & Accuracy(\%) \\ \midrule
Local          & 20.57                              & 127.76                              & 67.06        \\ 
FedStar & \textbf{30.95} & \textbf{265.01} & \textbf{67.48} \\ \bottomrule
\end{tabular}
\caption{Resource cost and performance comparison of local training (single-channel) and FedStar (decoupled dual-channel) in a cross-domain non-IID setting (\ie, BIO, SN, CV). We present the average FLOPs per client per round, local model size, and average test accuracy of each method.}
\label{tablefedstar}
}
\end{table}

To this end, an emerging area of research in FGL focuses on structural knowledge utilization. Among these studies, FedStar \cite{tan2023fedstar} achieves state-of-the-art accuracy across various datasets by employing a feature-structure decoupled dual-channel GNN architecture. 
This approach is promising as it isolates structural knowledge learning from features and facilitates its sharing in FGL.
However, FedStar's basic dual-channel design struggles with efficiency. 
While the additional channel enhances the learning of structural knowledge, it also significantly increases resource demands. 
Moreover, the simple decoupling design lacks interaction between the two channels, which may result in insufficient utilization of the knowledge each channel learns individually. This limitation can hinder their ability to accommodate different data distributions (\ie, non-IID) across clients, leading to persistently inconsistent updates and ultimately degrading the performance of the global model.

We then conduct preliminary experiments to better illustrate the above.
As shown in Table \ref{tablefedstar}, in a cross-domain non-IID setting, FedStar shows only a slight performance improvement (\ie, 0.42\%) over the local training with single-channel GNNs while incurring approximately an additional 10G FLOPs per client per round. This significant local computational demand poses a critical challenge for FGL, especially considering that many client devices (\eg, smartphones and tablets) have limited computational resources and participation time. Such demands may prevent clients from completing tasks, thereby severely impairing both local and global model performance.
Furthermore, FedStar lacks parameter efficiency.
The basic dual-channel architecture results in a local model size that is more than double that of single-channel networks, potentially leading to heavy network bandwidth and communication delays during model deployment, especially when the participating clients are enormous. These limitations indicate that the basic decoupling design is inefficient, and to address the non-IID data issues, the inherent structural knowledge across different domains still requires further exploration.

To overcome the aforementioned limitations,
we introduce FedDense, an FGL framework that optimizes the efficiency of structural knowledge utilization with dual-densely connected GNNs. To achieve comprehensive learning on diverse and underexploited structures, FedDense first introduces a structural vector that explicitly encodes the structural knowledge inherent within the graph itself. 
Furthermore, FedDense advances knowledge acquisition on top of the basic decoupled GNNs. Specifically, we introduce a Dual-Densely Connected (DDC) architecture, where the stacked GNN layers in both decoupled channels are densely connected with their feature maps, facilitating the multi-scale (\ie, one-hop to multi-hop) insights embedded in feature maps of both channels to be comprehensively tapped.
Finally, considering the resource constraint in FGL, we propose a very narrow layer design and implement a selective parameter sharing strategy within the DDC architecture to achieve high efficiency. 
As a result, each client in FedDense is capable of performing tasks with minimal resource demands. There are three key contributions of our work:
\begin{itemize}
\item[$\bullet$] In FedDense, we optimize the structural knowledge utilization within the graph data itself and the feature maps of each GNN layer, thereby mitigating non-IID data issues caused by diverse and underexploited graph structures.
\item[$\bullet$] By designing narrow layers and a selective parameter sharing strategy, FedDense ensures excellent performance while significantly reducing resource demands of model training, effectively addressing the efficiency concerns in FGL.
\item[$\bullet$] We conduct extensive experiments in four non-IID settings with 15 across 4 different domains, demonstrating that FedDense consistently outperforms baselines by a large margin in terms of test accuracy and convergence speed while requiring minimal computational demand and communication cost.
\end{itemize}

\section{Related Work}
\subsubsection{GNNs with Inherent Structure Knowledge. } 
Graph Neural Networks (GNNs) are a class of neural networks specifically designed to process and analyze graph-structured data \cite{zhou2020graph-survy}.
A fundamental characteristic of most GNNs is the message-passing mechanism, where each node representation is updated by aggregating information from its neighboring nodes' features \cite{message-passing1,message-passing2}. 
This mechanism enables GNNs to excel in tasks such as node classification, link prediction, and graph classification \cite{liu2023survey,zhou2020graph-survy}. 
However, recent studies increasingly recognize the fact that the feature-based message-passing mechanism falls short in differentiating and capturing the inherent structural knowledge of graph data \cite{failtostructureknowledge,failtostructureknowledge2}.
Most existing solutions aim to extract structural/positional encodings to explicitly represent structural information \cite{cui2022positionalstructembdings,positionemmbedding,positionICLR2022,limited-gnns-Representations-position}. Nevertheless, these approaches often overlook the additional insights that structural representations can provide at the feature map level, highlighting the need for further research in this area.

\subsubsection{Federated Graph Learning (FGL).}
FGL is an emerging field that allows GNNs to train on distributed graph data, thereby enhancing the potential of GNNs \cite{zhang2021FGL-survy}. 
A major challenge in FGL is the severe non-IID nature of graph data. Unlike typical Euclidean data, such as images, graphs are inherently more heterogeneous \cite{pan2023lumosgraphHeterogeneity}.
To address this challenge, \cite{xie2021GCFL} introduce a dynamic client clustering framework that reduces structural and feature heterogeneity within clusters, thereby improving learning efficiency and performance. \cite{wang2022graphfl} incorporate meta-learning techniques to manage non-IID graph data while maintaining generalizability. Additionally, \cite{tan2023fedstar} propose a feature-structure decoupled framework to extract and share structural information among graphs, enhancing the ability to capture structure-based domain-invariant knowledge.
However, while existing FGL methods have made breakthroughs in addressing non-IID issues, they often neglect considerations of resource consumption (\eg, communication bandwidth), which are essential for distributed paradigms. 
Therefore, ensuring the efficiency of FGL while addressing the non-IID problem remains a challenging and underdeveloped area in FGL research.

\section{Preliminaries}
\subsection{Graph Neural Networks (GNNs)}
A typical graph $G = \left(V, E\right)$ consists of a set of nodes $V$ and a set of edges $E$, where each node $v \in V$ is associated with a feature vector $\mathbf{x}_v$. We denote the representation of node $v$ as $\mathbf{h}_v$, and it can be iteratively updated by aggregating the representations of its one-hop neighbors $\mathcal{N}\left(v\right)$ as:
\begin{equation}
\mathbf{m}_v^{(\ell)}=\text{AGGREGATE}\left(\left\{\mathbf{h}_u^{(\ell-1)}|u\in\mathcal{N}(v)\right\}\right),
\end{equation}
\begin{equation}
\mathbf{h}_v^{(\ell)}=\text{UPDATE}\left(\mathbf{h}_v^{(\ell-1)},\mathbf{m}_v^{(\ell)}\right),
\end{equation}
where $\mathbf{h}_v^{(\ell)}$ is the updated representation of the node $v$ at the $\ell$-th layer. Different $\text{AGGREGATE}$ and $\text{UPDATE}$ functions allow for the implementation of various types of GNNs with distinct focuses \cite{zhou2020graph-GNNmethod}.

GNNs can be applied to various tasks, such as node classification, link prediction, and graph classification. 
In this paper, we focus primarily on graph classification, where GNNs combine the representations of all nodes to form a graph-level representation $\mathbf{h}_G$. This is typically achieved through pooling methods, such as average pooling, sum pooling, and max pooling. 

\subsection{Federated Graph Learning (FGL)}

A typical FGL system consists of a Parameter Server (PS) and a set of $N$ clients that collaboratively train a global GNN model. Each client $i$ holds a private graph dataset $d_i$, and the total samples across all clients are denoted as $D$. The training process of FGL is divided into $T$ rounds. At the start of each training round $t \in \{1, \ldots, T\}$, the PS distributes the global model parameters 
$\bar w^{(t)}$ to all clients.
Upon receiving $\bar{ w}^{(t)}$, each client $i$ performs local training on its private graph data $d_i$ and uploads the updated model parameters ${ w}_i^{(t)}$ back to the PS. At the end of round $t$, the PS aggregates these updates for the next round. The typical aggregation method used in FGL is FedAvg \cite{mcmahan2017Fedavg}, which averages the model updates from all clients by:
\begin{equation}
    \bar w^{(t+1)}=\sum_{i=1}^N\frac{|d_i|}{|D|} w^{(t)}_{i}, \label{fedavg}
\end{equation}
\noindent where $|d_i|$ denotes the size of data samples of client $i$ and $|D|$ represents the total size of samples over all clients.

The global model optimization in FGL aims to minimize the overall loss across all participating clients, denoted as:
\begin{equation}
\mathop{\arg\min}_{( w_1,w_2,\cdots, w_i)} \frac{1}{N} \sum_{i=1}^N \mathcal{L}_i\left(w_i\right),
\end{equation}
where $\mathcal{L}_i(\cdot)$ and $ w_i$ are the loss function and model parameters of client $i$, respectively.
However, due to the prevalence of non-IID data in real-life graph datasets, the performance of FGL is often suboptimal.

\begin{figure*}[t]
\centering
\includegraphics[width=1\textwidth]{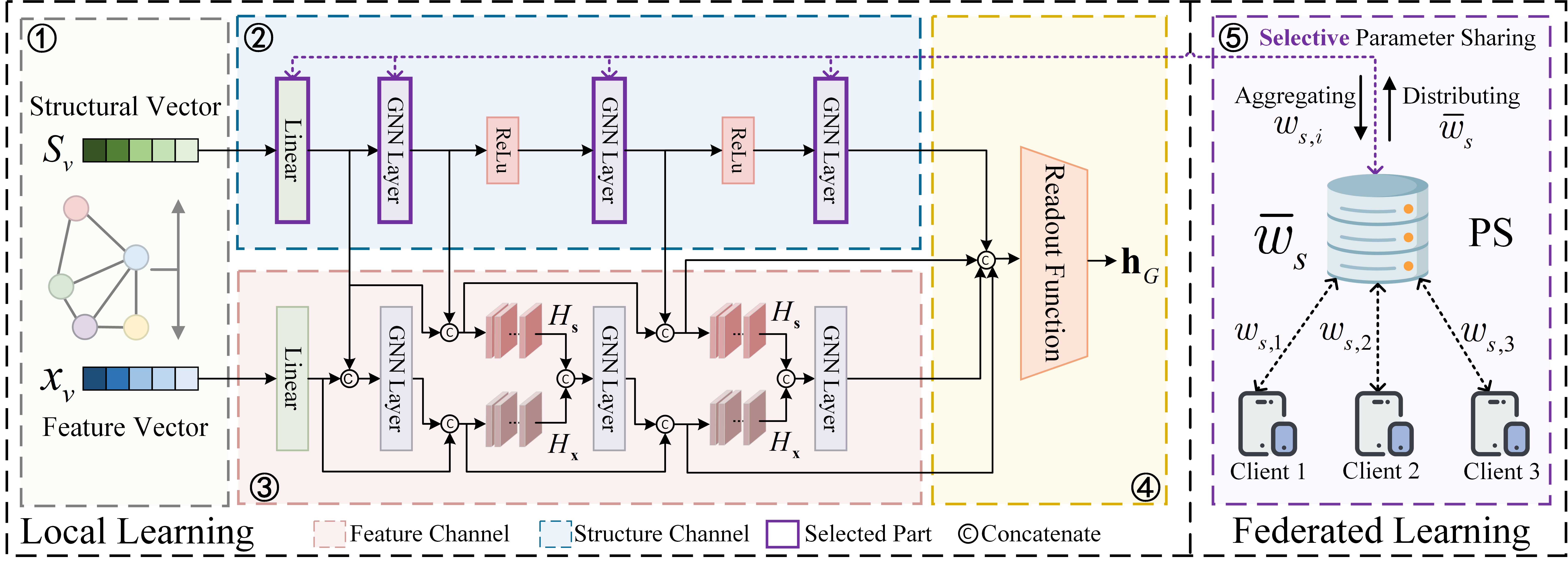} 
\caption{An overview of the proposed FedDense framework with 3 GNN layers as an example. The left box represents the local training process with structural patterns decoupling and DDC architecture of each client. The right box illustrates the global Selective Federated sharing scheme.}
\label{fig11}
\end{figure*}

\section{Methodology}
In this section, we detail our proposed FedDense framework, which is illustrated in Figure \ref{fig11}. FedDense focuses on better exploiting the inherent structural knowledge at two levels: the data level and the feature map level. 
At the data level, FedDense introduces a structural vector alongside node features (\ding{172}) to explicitly capture the unique structural patterns inherent in graph data itself. 
At the feature map level, we propose a Dual-Densely Connected GNN architecture. 
Specifically, we first employ dual-channel (\ding{173} and \ding{174}) GNNs to separately learn feature and structural knowledge with the decoupled vectors. 
Additionally, FedDense establishes dense connections between the dual-channel GNNs by integrating their feature maps at each layer in the feature channel (\ding{174}) and aggregates the collective knowledge within all hidden layers throughout the whole network to generate the final graph-level representation (\ding{175}). 
Furthermore, we design an efficient parameter sharing scheme that shares only the selected part of model parameters to achieve high communication efficiency (\ding{176}). 
Finally, we analyze the resource consumption of our framework and demonstrate that FedDense is highly resource-efficient with a significantly narrow layer design, achieving fewer model parameters, lower computational demand, and reduced communication costs.

\subsection{Structural Patterns Decoupling}
Existing GNNs primarily rely on feature information for message-passing. 
However, although GNNs implicitly incorporate structural knowledge through the message-passing mechanism by iteratively aggregating one-hop neighboring node features, this approach diminishes the direct learning of unique topological structures (\eg, node degree) in graph data. 
As demonstrated in Figure \ref{heatmap}, inherent structural patterns also carry significant and distinctive information, especially in cross-domain scenarios. Therefore, it is crucial to explicitly leverage structural knowledge and learn from both feature and structural information.

To this end, inspired by \cite{positionICLR2022,tan2023fedstar}, we introduce a structural vector into each graph node. 
Specifically, in addition to the node's feature vector $\mathbf{x}_v$, we extract and convert the node's underlying topological information into a structural vector $\mathbf{s}_v$, defined as: 
\begin{equation}
\mathbf{s}_v=f\left([s_1,s_2,s_3,\ldots,s_n]\right),
\end{equation}

\noindent where $[s_1,s_2,s_3,\ldots,s_n]$ are structural information encodings. To capture the local/global structural patterns of graph data, potential options include one-hot degree vectors, random walk transition matrices, and positional embeddings \cite{cui2022positionalstructembdings,positionICLR2022}. The function $f(\cdot)$ serves as a fusion function, which can be implemented using techniques such as concatenation, fully connected layers, or pooling \cite{Multi-Layers-Feature-Fusion}. Notably, both the structural encodings $[s_1,s_2,s_3,\ldots,s_n]$ and the function $f(\cdot)$ can be tailored according to the specific task, highlighting the versatility and adaptability of this approach.

Incorporating the structural vector adds an additional dimension of structural knowledge to each node, allowing the unique structural patterns embedded in the graph to be fully exploited. Consequently, the node representations become more robust and informative, combining both feature and structural information, denoted as $\{\mathbf{x}_v, \mathbf{s}_v\}$.


\subsection{Dual-Densely Connected Architecture}
\cite{tan2023fedstar} introduce a feature-structure decoupled dual-channel GNN architecture in FGL. This approach provides a solid starting point by allowing independent capture and processing of feature and structural information in graph data. 
However, simple decoupling has its limitations, as it overlooks internal interactions within the decoupled networks and the potential and informative multi-scale insights across their feature maps. 
A traditional GNN layer can be viewed as an aggregation from one-hop neighbors. Therefore, each layer of stacked GNNs is able to fetch different scales of local or global feature and structure insights within multi-hops, indicating that the feature maps in GNNs are highly informative, especially in a decoupled dual-channel architecture.

Inspired by this, we propose a novel Dual-Densely Connected (DDC) architecture. Specifically, we first employ dual-channel GNNs to separately learn feature and structural information with decoupled feature vector $\mathbf{x}_v$ and structural vector $\mathbf{s}_v$ at the data level. 
Additionally, we establish dense connections between the dual channels at the feature map level. Each layer in the feature channel receives additional inputs from the outputs of all preceding layers in both channels. This dual-dense connectivity allows the multi-scale insights of feature maps in both channels to be collectively leveraged.

\subsubsection{Initialization.} The DDC architecture employs two parallel GNNs: one for feature information and one for structural information. Both channels start with a linear initialization layer. In the feature channel, the feature vector $\mathbf{x}_v$ of each node $v$ is processed by a linear layer, transforming it into a hidden representation $\mathbf{x}_v^{(0)}$. Simultaneously, in the structural channel, the corresponding structural vector $\mathbf{s}_v$ is passed through a separate linear layer, producing the representation $\mathbf{s}_v^{(0)}$ with the same dimension as $\mathbf{x}_v^{(0)}$.

\subsubsection{Dual-Dense Connectivity. } After initialization, both channels follow $L$ stacked GNN layers. To maintain simplicity and preserve the integrity of structural information, the input to the $\ell$-th GNN layer in the structural channel is directly derived from the output of the previous layer $\mathbf{s}_v^{(\ell-1)}$ in this channel, where $\ell \in \{1,\cdots, L\}$. Meanwhile, to enhance internal interactions within the decoupled networks and leverage multi-scale information from their stacked hidden layers, each layer in the feature channel 
receives the feature maps of all preceding layers from both channels as input:
\begin{equation}\mathbf{c}_v^{(\ell)}=\textrm{Concat}[\boldsymbol{\alpha}_v^{(\ell)},\boldsymbol{\beta}_v^{(\ell)}]
,\end{equation}
\noindent where $\textrm{Concat}\left[\cdot\right]$ denotes the concatenation operation. The $\boldsymbol{\alpha}_v^{(\ell)}$ and $\boldsymbol{\beta}_v^{(\ell)}$ can be denoted as:
\begin{equation}\boldsymbol{\alpha}_v^{(\ell)}=H_\mathbf{x}(\textrm{Concat}[\mathbf{x}_v^{(0)},\mathbf{x}_v^{(1)},\cdots,\mathbf{x}_v^{(\ell - 1)}]),
\end{equation}
\begin{equation}\boldsymbol{\beta}_v^{(\ell)}=H_\mathbf{s}(\textrm{Concat}[\mathbf{s}_v^{(0)},\mathbf{s}_v^{(1)},\cdots,\mathbf{s}_v^{(\ell - 1)}]),
\end{equation}
\noindent where $\mathbf{x}_v^{(0)}, \mathbf{x}_v^{(1)}, \cdots, \mathbf{x}_v^{(\ell - 1)}$ and $\mathbf{s}_v^{(0)}, \mathbf{s}_v^{(1)}, \cdots, \mathbf{s}_v^{(\ell - 1)}$ refer to the feature maps produced in layers $0$ through $\ell-1$ from the feature and structural channels, respectively. We define $H_\mathbf{x}(\cdot)$ and $H_\mathbf{s}(\cdot)$ as non-linear transformations between hidden GNN layers in the feature channel, consisting of a composite function of operations such as Batch Normalization (BN), Dropout, rectified linear units (ReLU), and Pooling.


For the final graph-level embedding $\mathbf{h}_G$, rather than relying solely on the output of the final layer, we consider and concatenate the feature maps of all the hidden layer outputs generated across both channels. The concatenated representation is then transformed into the graph-level embedding via a readout function. 

With the DDC architecture, FedDense not only ensures the independent learning of structural knowledge with two parallel channels but also guarantees the different insights they learn individually are fully integrated and leveraged with dual-dense Connectivity. Additionally, by combining all feature maps throughout the network to generate the final graph embeddings, FedDense considers the collective knowledge across both channels, thus achieving robust and comprehensive learning of graph data.

\subsection{Selective Federated Sharing}
Unlike traditional FGL methods, where all model parameters are shared among clients, FedDense restricts parameter sharing to the structural parameters only, specifically the learnable parameters of each layer in the structural channel. Hence, Eq. (\ref{fedavg}) can be reformulated as follows:

\begin{equation}
    \bar w^{(t+1)}_{s}=\sum_{i=1}^N\frac{|d_i|}{|D|} w^{(t)}_{s,i}
\end{equation}

\noindent where $\bar w^{(t+1)}_{s}$ represents the aggregated structural parameters at the PS, and $ w^{(t)}_{s,i}$ denotes the updated structural parameters of client $i$ in round $t$. The feature parameters are neither shared nor updated through federated learning but are instead optimized locally within each client.

This approach is adopted for three key reasons: (1) Communication Efficiency. Transmitting the full set of parameters from both the structure and feature channels can lead to significant communication delays and increased data transfer costs, especially in environments with constrained bandwidth. Therefore, selecting and transmitting only a subset of parameters is essential for maintaining efficiency. (2) The Significance of Structural Information. Structural information is crucial as it provides deep insights into the underlying topology of a graph. 
Elements such as node degree, connectivity, and overall graph structure encapsulate essential characteristics that are often more informative than feature attributes alone, especially in heterogeneous or cross-domain graphs. 
In scenarios where communication bandwidth is limited, prioritizing the transmission of structural parameters becomes advantageous. (3) Synergy enhancements via dual-dense connectivity. The dense integration of feature and structural channels at the feature map level enables synergistic interaction between these two types of information in the feature channel. Even though feature training is conducted locally, the structural knowledge shared through federated learning can significantly benefit feature learning. This dual-dense connectivity in the feature channel ensures that local feature updates are informed by the global structural context, leading to more robust and comprehensive training. Considering these factors, FedDense shares only the structural parameters to achieve reduced communication costs while preserving essential information. 

\begin{table*}[ht]
{\fontsize{10}{\baselineskip}\selectfont
\begin{tabular}{@{}cc@{\hskip 0.1cm}ccc@{\hskip 0.1cm}ccc@{\hskip 0.1cm}ccc@{\hskip 0.1cm}cc@{}}
\toprule
Settings (\# domains) &  & \multicolumn{2}{c}{Single (1)}          &  & \multicolumn{2}{c}{Cross-Sim (2)}       &  & \multicolumn{2}{c}{Cross-Diff (3)}      &  & \multicolumn{2}{c}{Multi (4)}           \\ \cmidrule(r){1-1} \cmidrule(lr){3-4} \cmidrule(lr){6-7} \cmidrule(lr){9-10} \cmidrule(l){12-13} 
\# datasets          &  & \multicolumn{2}{c}{7}               &  & \multicolumn{2}{c}{9}               &  & \multicolumn{2}{c}{8}               &  & \multicolumn{2}{c}{15}              \\ \cmidrule(r){1-1} \cmidrule(lr){3-4} \cmidrule(lr){6-7} \cmidrule(lr){9-10} \cmidrule(l){12-13} 
Metrics              &  & avg.acc             & avg.gain      &  & avg.acc             & avg.gain      &  & avg.acc             & avg.gain      &  & avg.acc             & avg.gain      \\ \midrule
Local                &  & 75.50±1.77          & -             &  & 72.12±1.81          & -             &  & 67.06±1.84          & -             &  & 71.12±1.14          & -             \\ \midrule
FedAvg               &  & 75.07±3.72          & -0.43         &  & 71.73±1.01          & -0.39         &  & 64.80±2.60          & -2.26         &  & 68.67±0.06          & -2.45         \\
FedProx              &  & 74.00±2.03          & -1.50         &  & 72.57±1.82          & 0.45          &  & 63.94±3.64          & -3.12         &  & 69.18±0.08          & -1.94         \\
GCFL                 &  & 75.36±1.17          & -0.14         &  & 72.98±1.37          & 0.86          &  & 64.54±1.88          & -2.25         &  & 70.32±0.10          & -0.8          \\
FedStar              &  & 79.32±2.47          & 3.82          &  & 74.42±2.37          & 2.30           &  & 67.48±3.04          & 0.42          &  & 73.70±1.54          & 2.58          \\ \midrule
FedDense ($r$ = 16)     &  & \underline{79.86}±2.62 & \underline{4.36} &  & \underline{74.51}±2.35 & \underline{2.39} &  & \underline{71.66}±2.03 & \underline{4.60} &  & \underline{74.35}±1.85 & \underline{3.23} \\
FedDense ($r$ = 32)     &  & \textbf{80.16}±2.74 & \textbf{4.66} &  & \textbf{75.74}±2.44 & \textbf{3.62} &  & \textbf{72.44}±2.05 & \textbf{5.38} &  & \textbf{75.31}±1.68 & \textbf{4.19} \\ \bottomrule
\end{tabular}
}
\caption{Performance in four non-IID settings. We present the average accuracy and gain over Local for all FGL methods. The best performances in each setting are highlighted in bold, while the second-best performances are underlined.}
\label{tabel1111}
\end{table*}

\subsection{Analysis}
We assume that the primary resource bottlenecks of FedDense arise from the GNN layers within dual-densely connected channels and take GCN \cite{Kipf2016SemiSupervisedCW-GCN} as an analysis example. 
For a given graph $G = \left(V, E\right)$, we denote $\left|V\right|$ and $|E|$ as the total number of nodes and edges. 
The dimensions of each input and output feature map for each GCN layer are denoted by $a$ and $b$. 
We denote the time complexity as $\Theta $ and the number of parameters as $\left|\theta\right|$.
Generally, the layer parameters are predominantly determined by its weight matrix $\mathbf{W}\in\mathbb{R}^{a\times b}$. 
Therefore, for each GCN layer, we can derive that $\Theta =O(|E|a + |V|ab)$ and $\left|\theta\right| = a\times b$ \cite{wu2020comprehensive-GCNbigO}, respectively. In FedDense, if the output size for all layers is set to the same value, denoted by the hyperparameter $r$, it follows that the $\ell$-th GNN layer has $2\times\ell\times r$ and $r$ input feature maps in feature and structural channels, respectively. In this way,  $\Theta $ and $\left|\theta\right|$ of the $\ell$-th layer can be represented as $O(|E|2\ell r + |V| 2\ell r^2)$ and $2\ell r^2$ for feature channel, and $O(|E|r + |V|r^2)$ and $r^2$ for structural channel. It is noteworthy that GNNs often achieve optimal performance with shallow architectures, where $\ell$ in GNNs is relatively small, typically between 2 and 3 \cite{zhang2021evaluatingGNNSDepth}, and thus can be treated as a constant when analysis. Therefore, $\Theta $ and $\left|\theta\right|$ of the $\ell$-th GCN layer in both channels can be summed to $O(|E|r + |V|r^2)$ and  $2\ell r^2+r^2$, respectively. It is evident that for a given graph $G$, both  $\Theta $ and $\left|\theta\right|$ of FedDense are significantly reduced as $r$ decreases. Additionally, since the shared parameters in FedDense are limited to the structural channel only, the communication cost per round remains minimal due to the reduced model parameters.

According to the above analysis, it is evident that for a given graph dataset, limiting the local model in FedDense to a narrow layer design (\eg, $r$ = 16 or $r$ = 10) significantly reduces the resource demand in both local training and parameter sharing. Furthermore, thanks to our DDC architecture, even with a very narrow layer design, FedDense maintains efficient and comprehensive acquisition of both local and global knowledge. As shown in the next section, with a significantly small $r$ compared to all baselines, FedDense achieves excellent results on the test datasets while minimizing computational demands and communication costs.

\section{Experiments}\label{experiments}
\subsection{Datasets and Experimental Setup}

\subsubsection{Datasets.}
We utilize a total of 15 datasets \cite{morris2020tudataset} across 4 different domains: seven Molecules datasets (MUTAG, BZR, COX2, DHFR, PTC-MR, AIDS, NCI1), two Bioinformatics datasets (ENZYMES, DD), three Social Networks datasets (COLLAB, IMDB-BINARY, IMDB-MULTI), and three Computer Vision datasets (Letter-low, Letter-high, Letter-med). 
To simulate data heterogeneity in FGL, we establish four different non-IID settings: (1) a single-domain setting (\ie, Single) using only the Molecules datasets; (2) a cross-domain setting (\ie, Cross-Sim) using datasets from similar domains (Molecules and Bioinformatics); (3) another cross-domain setting (\ie, Cross-Diff) utilizing datasets from completely different domains (Bioinformatics, Social Networks, and Computer Vision); and (4) a multi-domain setting (\ie, Multi) incorporating datasets from all four domains. In each setting, the graph data for each client is derived from one of the corresponding datasets and is randomly split into a ratio of 8:1:1 for training, validation, and testing.
\begin{figure}[t]\centering
    \begin{subfigure}[t]{0.495\linewidth}\centering
                \label{Stability of clustering-line-1}
                \includegraphics[width=1.0\linewidth]{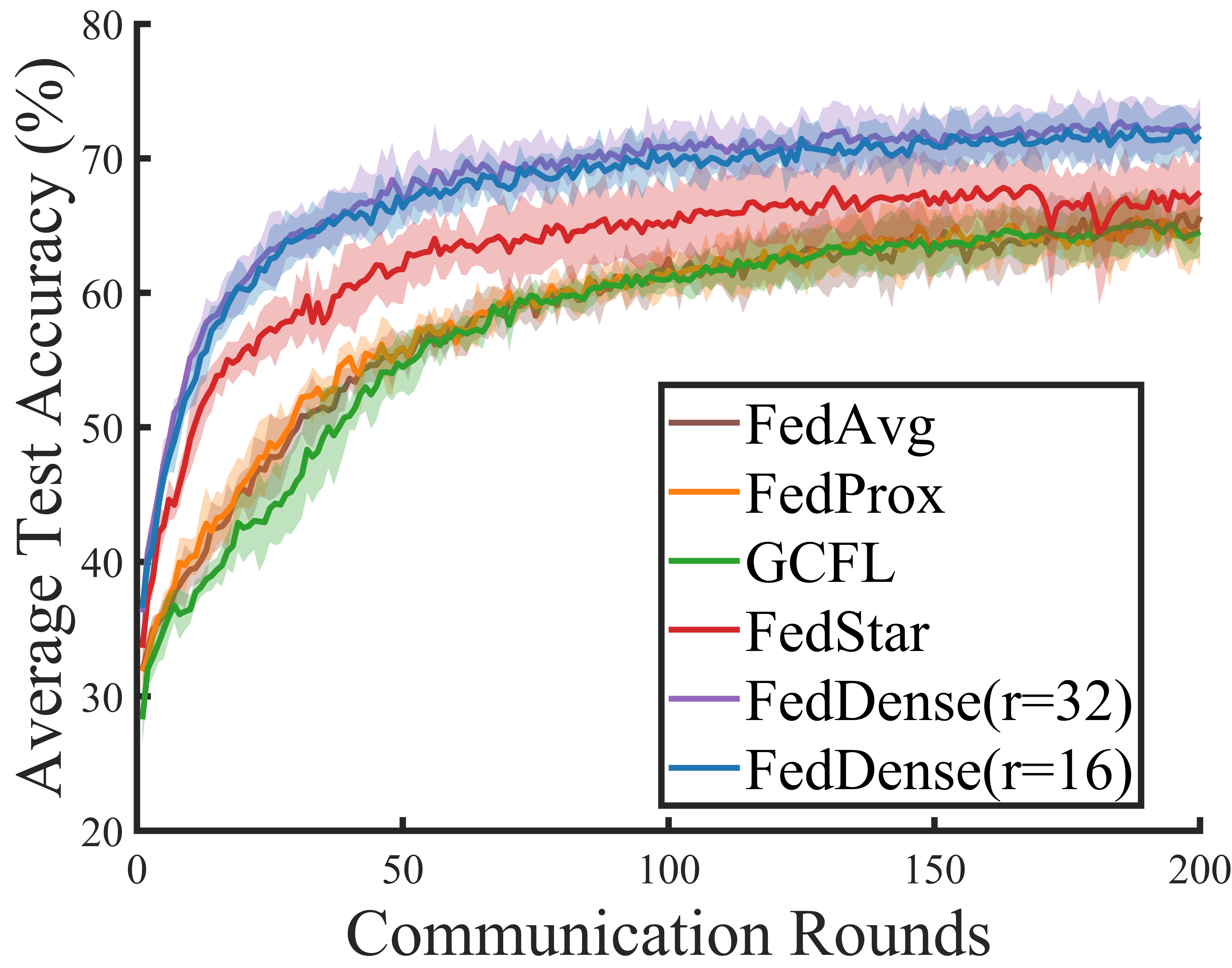}
                \subcaption{Cross-Diff}
    \end{subfigure}
    \begin{subfigure}[t]{0.495\linewidth}\centering
                \label{Stability of clustering-line-2}
                \includegraphics[width=1.0\linewidth]{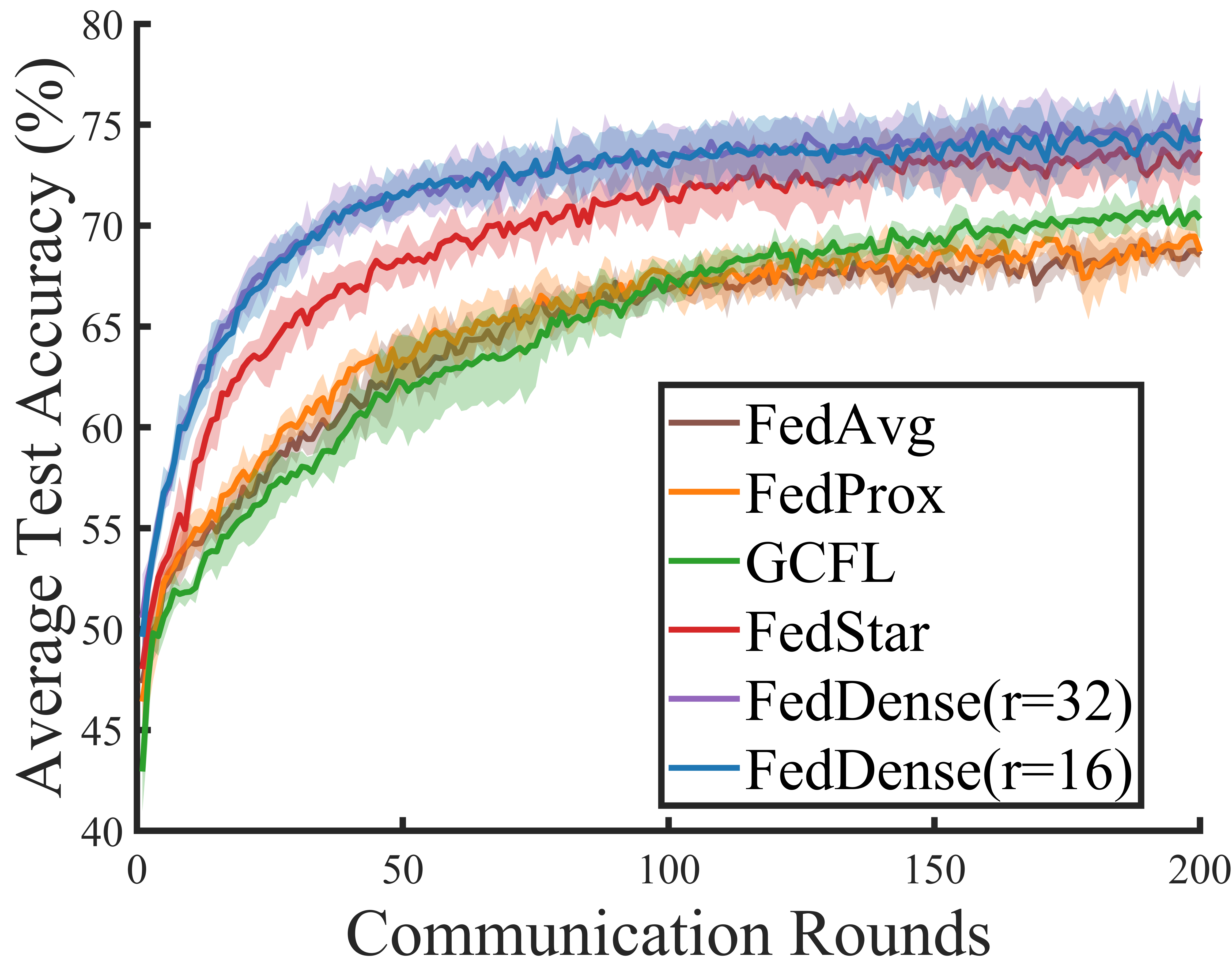}
                \subcaption{Mulit}
            
    \end{subfigure}
    
   \caption{Test accuracy curves of FedDense and four baselines along the communication rounds under highly heterogeneous conditions.}\label{converge}
\end{figure}
\begin{figure*}[t]\centering
    \begin{subfigure}[t]{0.246\textwidth}\centering
                \label{fig:time-to-accuracy-end-to-end-cifar10}
                \includegraphics[width=1.0\linewidth]{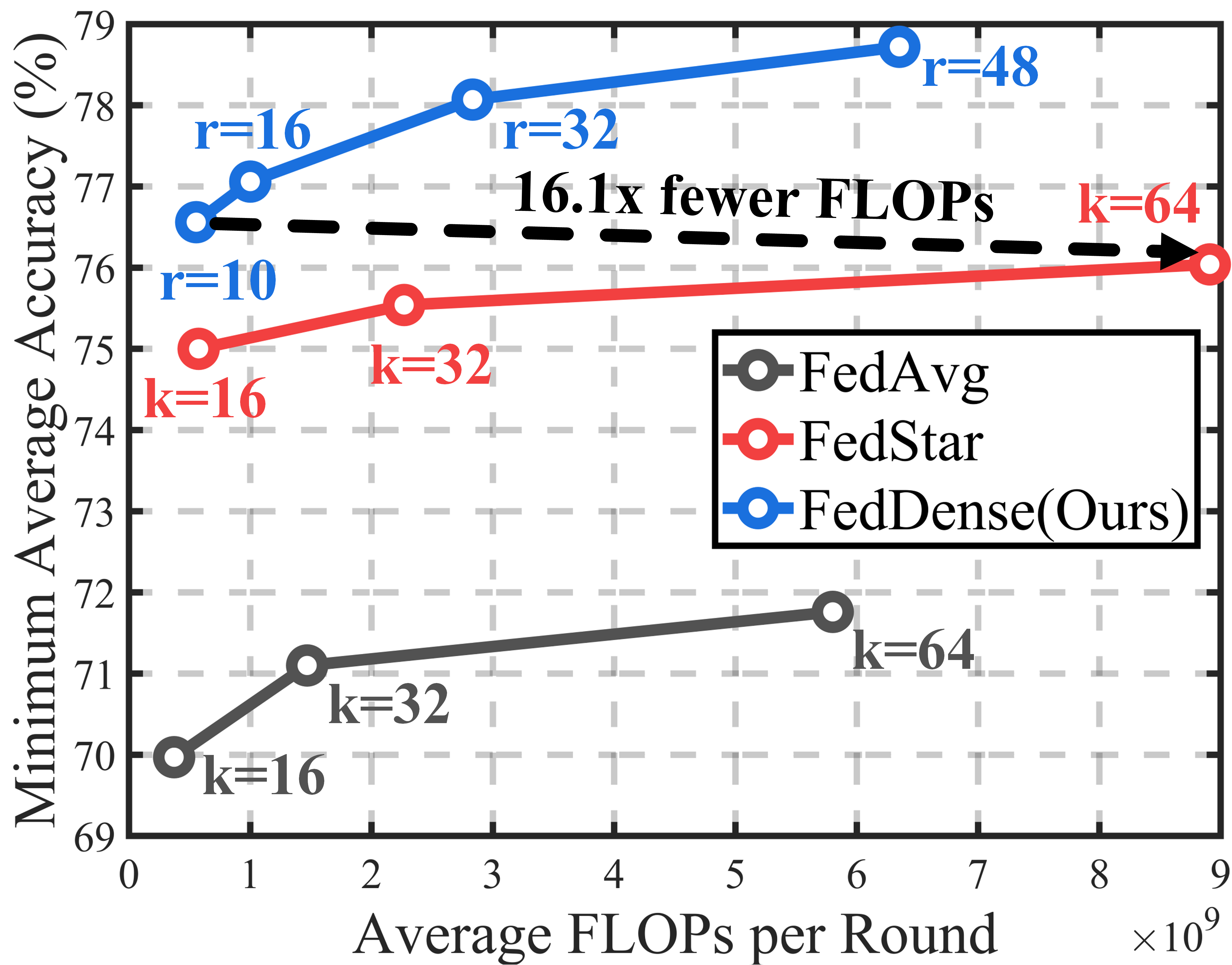}
                \subcaption{Single}
            
    \end{subfigure}
    \begin{subfigure}[t]{0.246\textwidth}\centering
                \label{fig:time-to-accuracy-end-to-end-cifar100}
                \includegraphics[width=1.0\linewidth]{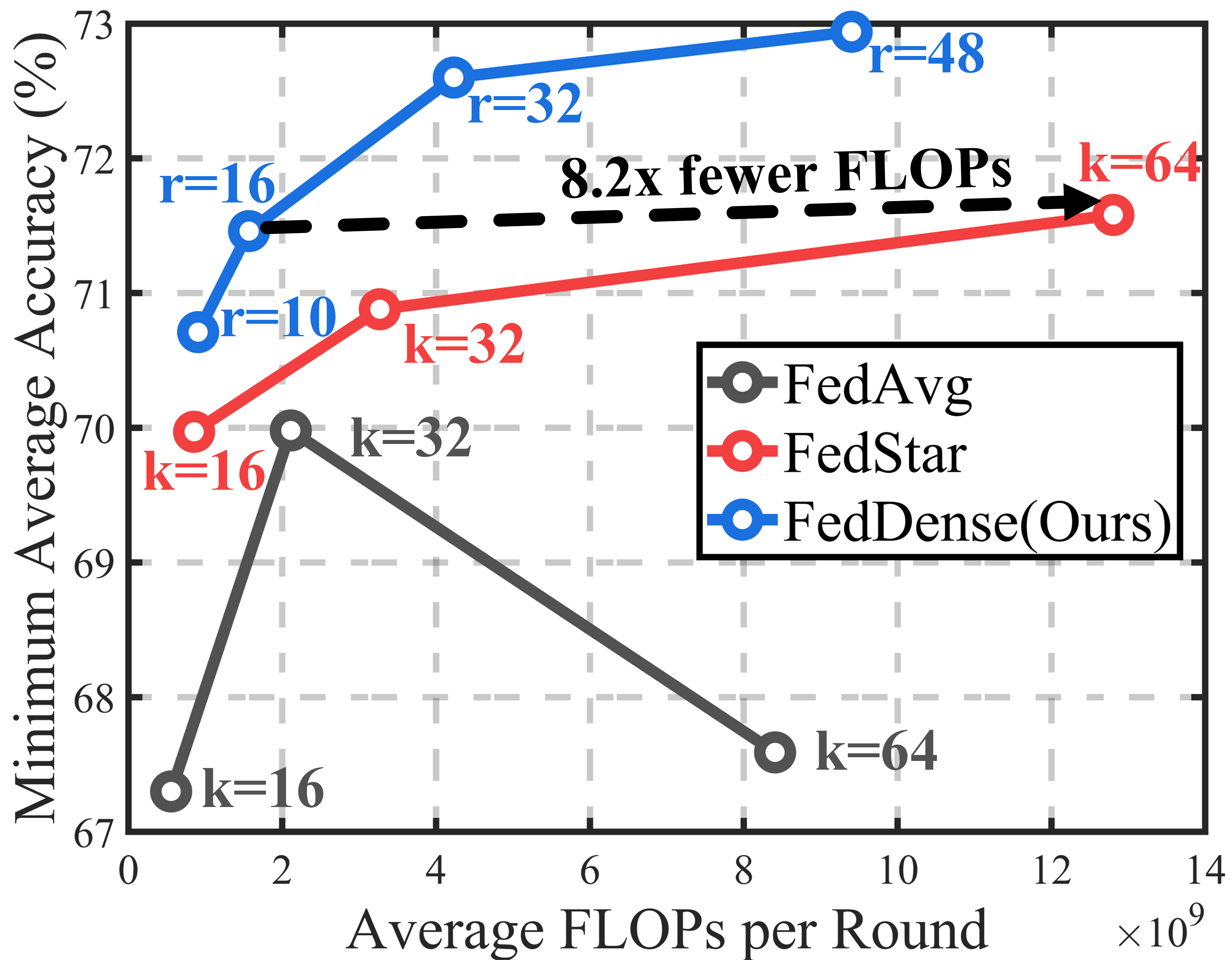}
                \subcaption{Cross-Sim}
    \end{subfigure}
    \begin{subfigure}[t]{0.246\textwidth}\centering
                \label{fig:time-to-accuracy-end-to-end-googlespeech}
                \includegraphics[width=1.0\linewidth]{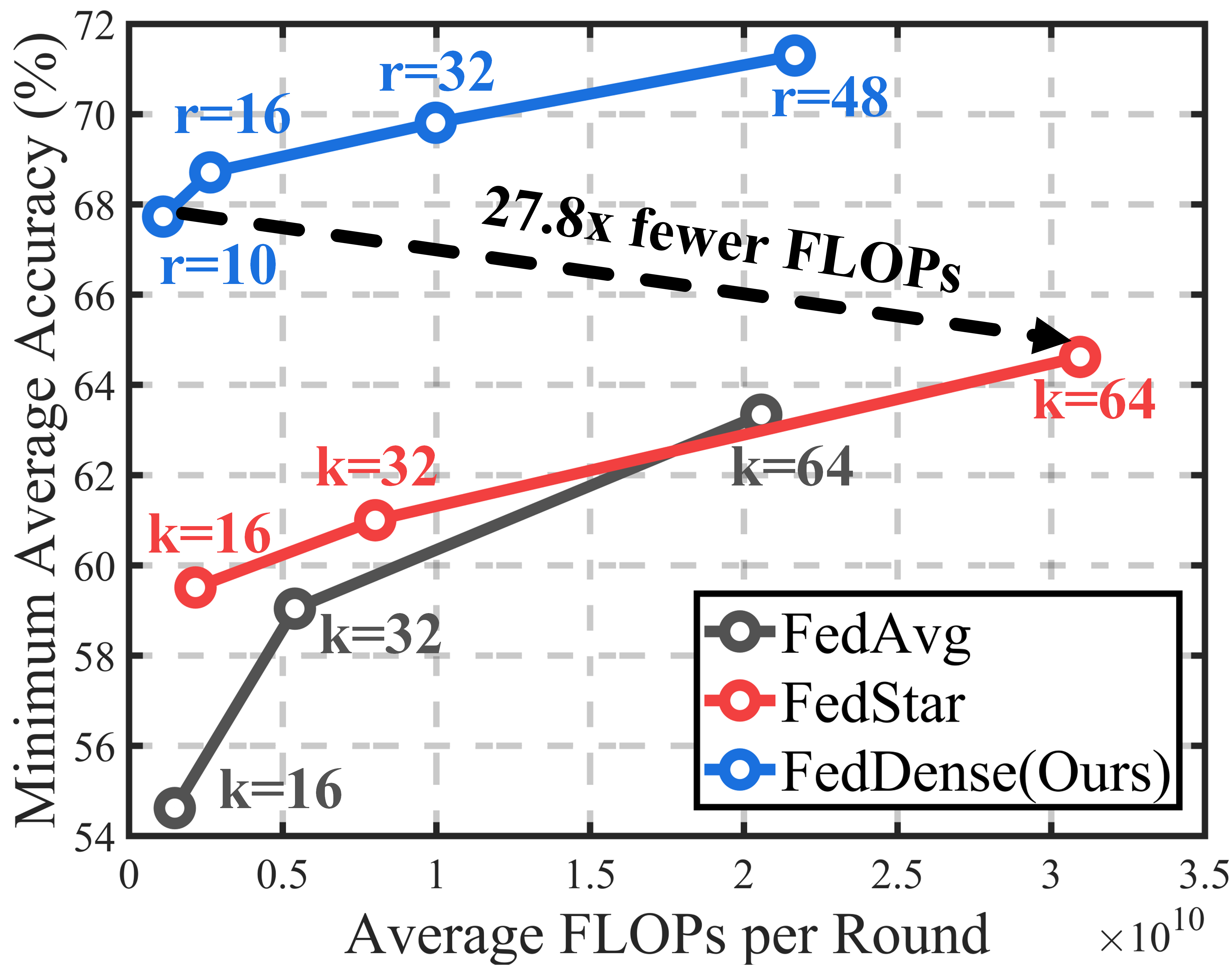}
                \subcaption{Cross-Diff}
    \end{subfigure}
    \begin{subfigure}[t]{0.246\textwidth}\centering
                \label{fig:time-to-accuracy-end-to-end-avazu}
                \includegraphics[width=1.0\linewidth]{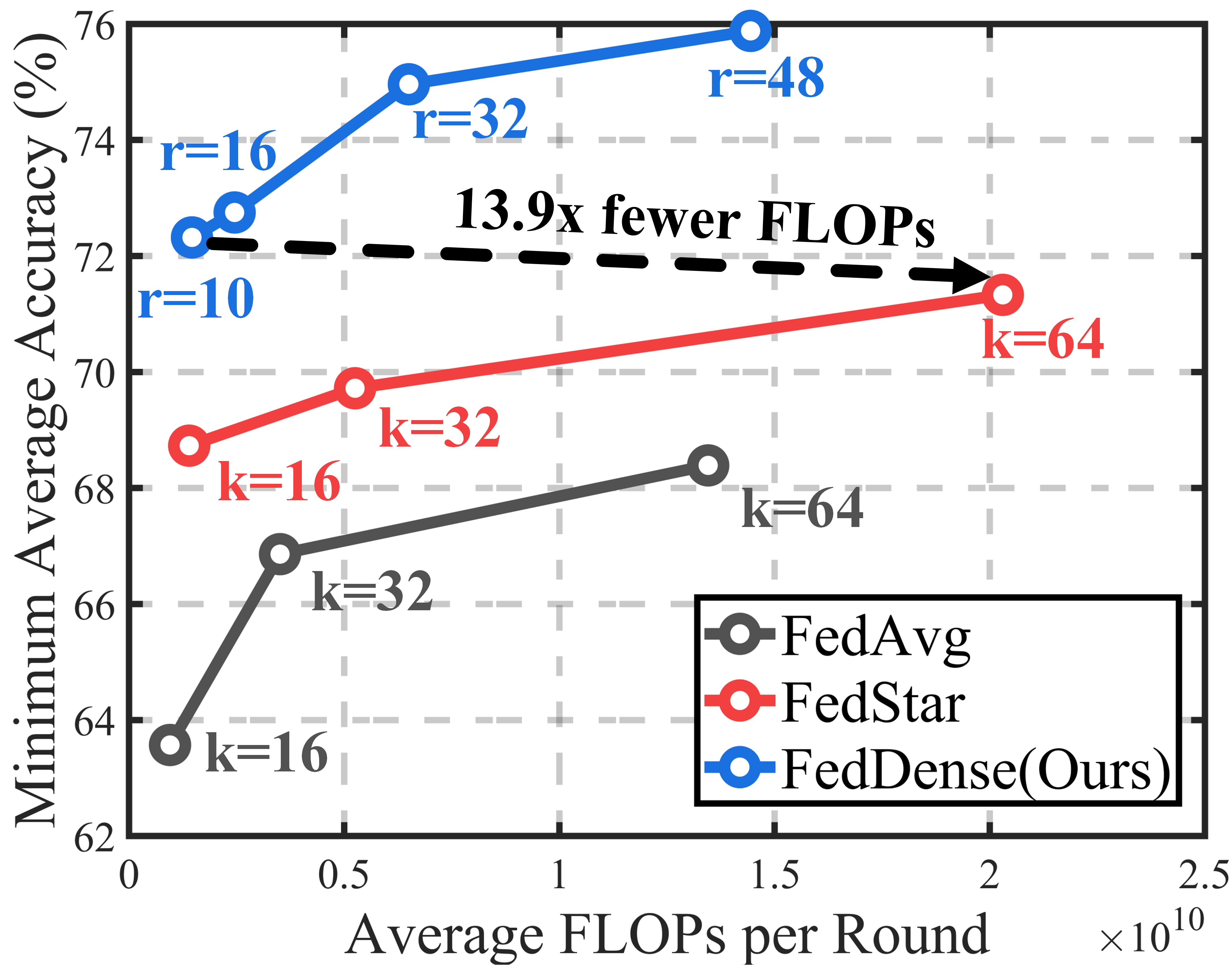}
                \subcaption{Mulit}
    \end{subfigure}
    \caption{Computation efficiency in four non-IID settings, presenting the minimum average accuracy across five random repetitions and the average FLOPs per client per round for each FGL method. $r$ and $k$ denote the output size of the local GNNs in FedDense and other FGL methods, respectively.}\label{fig:time-to-accuracy-end-to-end}\label{compution}
\end{figure*}
\subsubsection{Baselines.}We employ five baselines in our experiments: (1) Local, where each client conducts model training locally without any communication with others; (2) FedAvg \cite{mcmahan2017Fedavg}, a standard FGL approach that aggregates client models by averaging their local updates; (3) FedProx \cite{MLSYS2020_1f5fe839fedproxy}, where a regularization term in the loss function was proposed to handle system and statistical heterogeneity; (4) GCFL \cite{xie2021GCFL}, which tackles non-IID graph data through a dynamic clustering technique based on GNN gradients to group clients with similar data distributions; and (5) FedStar \cite{tan2023fedstar}, a state-of-the-art FGL framework that decouples structural and feature learning and sharing across diverse graph domains.

\subsubsection{Implementation Details.}
To construct the structural vector, we align with the settings used in FedStar \cite{tan2023fedstar}. Specifically, we concatenate two types of structural encodings: a degree-based embedding representing vertex degrees with one-hot encoding and a random walk-based positional embedding which is computed based on the random walk diffusion process \cite{positionICLR2022,li2020distancegraphembeddings-RWPEcited}, both with dimensions of 16. The non-linear transformations in FedDense, consistent with other methods, apply ReLU followed by Dropout.
We utilize a 3-layer GIN \cite{xu2018powerful-GIN} in the feature channel for all methods and a 3-layer GCN \cite{Kipf2016SemiSupervisedCW-GCN} in the structural channel for FedDense and Fedstar.  We set the hidden size to 64 for all baselines. 
For FedDense, the hidden size of each layer is controlled by the hyperparameter $r$. We use a batch size of 128 and the Adam optimizer \cite{Kingma_Ba_2014adam} with a learning rate of 0.001 and a weight decay of $5\times10^{-4}$. 
The local epoch is set to 1, and the number of communication rounds is 200 for all FGL methods. 
All experiments are conducted on one NVIDIA GeForce RTX 4090 GPU and run for five random repetitions. 
More implementation details can be found in the Appendix.

\subsection{Experimental Results.}
\subsubsection{Accuracy Performance.}
As shown in Table \ref{tabel1111}, FedDense surpasses all competing baselines in four non-IID settings. 
In Cross-Diff and Multi settings, where the data across clients is more heterogeneous, all baselines exhibit severe performance degradation, with most methods failing to surpass the Local baseline. 
However, under these highly heterogeneous conditions, FedDense ($r$ = 32) achieves impressive average accuracy gains of 5.38\% and 4.19\%, respectively, significantly outperforming the existing state-of-the-art FedStar by notable margins of 4.98\% and 1.61\%. Remarkably, even with a small $r$ (\ie,$ r$ = 16), our framework still achieves excellent performance gain(\ie, 4.60\% and 3.23\%) and continues to surpass FedStar (\ie, 4.18\% and 0.65\%). 
The superior performance of FedDense can be attributed to its structural vector and DDC architecture. 
The additional dimension of structural knowledge and integration of both feature and structural insights significantly improve the knowledge acquisition across clients, thereby enhancing FedDense to model complex and diverse structural patterns in both local and cross-domain graphs.

\subsubsection{Convergence Analysis.}
Figure \ref{converge} illustrates the average test accuracy with standard deviation curves during training across five random runs for all methods.
In the Cross-Diff and Multi settings, where client data exhibits higher heterogeneity, FedDense consistently outperforms other methods in terms of average test accuracy and convergence speed.
For instance, in the Cross-Diff setting, FedDense reaches 65\% test accuracy by round 36 ($r$ = 32) and 38 ($r$ = 16), while FedAvg, FedProx, GCFL, and FedStar require 180, 165, 189, and 88 rounds, respectively, to reach 65\% test accuracy. 
The significant improvement, especially over FedStar, demonstrates that our DDC architecture greatly reinforces the federated structure knowledge acquisition at each training round, thus greatly leveraging the advantages of knowledge sharing in FGL and speeding up the convergence.

\subsubsection{Communication Cost.}
Table \ref{tab:my-table} presents the communication cost for FedAvg, FedStar, and FedDense. We divide communication cost into two parts: the parameter sharing payload per client per round during the FGL process and the size of the distributed local model during the deployment. 
Obviously, FedDense with $r$=16 only takes 14.7\% of the payload relative to the standard FGL paradigm, FedAvg, while maintaining a nearly equivalent model size. 
Furthermore, FedDense significantly reduces the communication payload by approximately 74.7\% compared to FedStar, with the model size being nearly 50.6\% smaller. 
Considering the outstanding performance of FedDense in terms of accuracy and convergence speed, these results indicate that the narrow layer design and selective parameter sharing in FedDense significantly reduce communication costs while successfully guaranteeing that essential and informative knowledge is effectively learned both locally and globally, thereby endowing FedDense with high communication efficiency throughout the entire FGL training process.

\begin{table}[t]
\centering
{\fontsize{9}{\baselineskip}\selectfont
\renewcommand{\arraystretch}{0.9}
\begin{tabular}{@{}ccc@{}}
\toprule
Comm. Cost &   Payload (KB) & Model Size (KB) \\ \midrule
FedAvg             & 97.70 ($\times$1.000)                         & 127.76 ($\times$1.000)         \\
FedStar            & 57.00 ($\times$0.583)                        & 265.01 ($\times$2.074)         \\
FedDense ($r$=32)     & \textbf{24.63 ($\times$0.252)}                & 219.88 ($\times$1.721)         \\
FedDense ($r$=16)     & \textbf{14.44 ($\times$0.147)}                & \textbf{134.32 ($\times$1.051)} \\ \bottomrule
\end{tabular}
}\caption{Communication cost (Comm. Cost) with relative ratios to FedAvg, covering both the parameter sharing payload per client per round and local model size for each client.}
\label{tab:my-table}
\end{table}

\subsubsection{Computational Efficiency.}
One of the primary advantages of our proposed framework is its remarkable computational efficiency. As illustrated in Figure \ref{compution}, FedDense significantly outperforms FedAvg and FedStar in terms of minimum average accuracy across five random repetitions while requiring minimal local computation. 
Although FedStar achieves better accuracy compared to FedAvg, its basic dual-channel GNN architecture introduces significant additional local computation demands, making it less suitable for distributed paradigms like FGL, especially given the often limited computational resources and participation time of each client. In contrast, FedDense stands out for its exceptional efficiency, achieving the best accuracy performance while demanding minimal computational resources in all non-IID settings. Notably, in the highly heterogeneous Cross-Diff setting, FedDense ($r$ = 10) surpasses FedStar ($k$ = 64) by a large margin in accuracy while requiring 27.8 times lower FLOPs per client per round, demonstrating that extracting knowledge from feature maps in each GNN layer significantly enriches both local and global knowledge acquisition in FGL. Despite constraints on computational resources and limited client participation time, FedDense delivers outstanding performance.

\section{Conclusion}
This paper proposes an effective framework, FedDense, to optimize federated graph learning with inherent structural knowledge and
dual-densely connected GNNs. To better exploit diverse structures, we decouple structural patterns at the data level and employ a dual-densely connected architecture at the feature map level. Moreover, we design narrow layers and adopt a selective parameter sharing strategy for high resource efficiency. The extensive experimental results demonstrate that FedDense can achieve state-of-the-art performance with minimum resource demands.

\bigskip

\bibliography{ref}
\newpage
\appendix
\onecolumn
\section{Appendix}
\subsection{Experimental Details}
In this appendix, we provide detailed descriptions of the experimental setups and specific configurations that were not fully elaborated in the main text due to space limitations.
\subsubsection{Data Splitting Details.} 
In this section, we detail the configuration of the data splitting across the four non-IID settings described in the main text. Specifically, For each non-IID setting—Single, Cross-Sim, Cross-Diff, and Multi—we systematically assigned datasets to clients and implemented a random split of 8:1:1 for training, validation, and testing. The precise configurations for each non-IID scenario are summarized in the tables below.
\begin{table}[h]
\centering
\caption{Detailed statistics of non-IID setting \textbf{Single}
}
\label{tab:my-table}
\resizebox{\textwidth}{!}{\begin{tabular}{@{}c|ccccccc@{}}
\toprule
Setting                              & \multicolumn{7}{c}{Single}                         \\ \midrule
Domain                               & \multicolumn{7}{c}{Chemical Molecules}             \\ \midrule
Datasets                             & MUTAG & BZR & COX2 & DHFRR & PTRC-MR & AIDS & NCI1 \\ \midrule
\# of clients                        & 1     & 1   & 1    & 1     & 1       & 1    & 1    \\
\# of graphs for training per client & 150   & 324 & 373  & 604   & 275     & 1600 & 3288 \\ \bottomrule
\end{tabular}}
\end{table}

\begin{table}[h]
\centering
\caption{Detailed statistics of non-IID setting \textbf{Cross-sim}
}
\label{tab:my-table}
\resizebox{\textwidth}{!}{\begin{tabular}{@{}c|ccccccccc@{}}
\toprule
Setting                              & \multicolumn{9}{c}{Cross-sim}                                                                                \\ \midrule
Domain                               & \multicolumn{7}{c|}{Chemical Molecules}                                 & \multicolumn{2}{c}{Bioinformatics} \\ \midrule
Datasets                             & MUTAG & BZR & COX2 & DHFRR & PTRC-MR & AIDS & \multicolumn{1}{c|}{NCI1} & ENZYMES            & DD            \\ \midrule
\# of clients                        & 1     & 1   & 1    & 1     & 1       & 1    & \multicolumn{1}{c|}{1}    & 1                  & 1             \\
\# of graphs for training per client & 150   & 324 & 373  & 604   & 275     & 1600 & \multicolumn{1}{c|}{3288} & 480                & 942           \\ \bottomrule
\end{tabular}
}
\end{table}

\begin{table}[h]
\centering
\caption{Detailed statistics of non-IID setting \textbf{Cross-diff}
}
\label{tab:my-table}
\resizebox{\textwidth}{!}{%
\begin{tabular}{@{}c|cccccccc@{}}
\toprule
Setting                              & \multicolumn{8}{c}{Cross-diff}                                                                                                       \\ \midrule
Domain                               & \multicolumn{2}{c|}{Bioinformatics} & \multicolumn{3}{c|}{Social Networks}                   & \multicolumn{3}{c}{Computer Vision}   \\ \midrule
Datasets                             & ENZYMES  & \multicolumn{1}{c|}{DD}  & COLLAB & IMDB-BINARY & \multicolumn{1}{c|}{IMDB-MULTI} & Letter-low & Letter-high & Letter-med \\ \midrule
\# of clients                        & 1        & \multicolumn{1}{c|}{1}   & 1      & 1           & \multicolumn{1}{c|}{1}          & 1          & 1           & 1          \\
\# of graphs for training per client & 480      & \multicolumn{1}{c|}{942} & 4000   & 800         & \multicolumn{1}{c|}{1200}       & 1800       & 1800        & 1800        \\ \bottomrule
\end{tabular}%
}
\end{table}

\begin{table}[H]
\centering
\caption{Detailed statistics of non-IID setting \textbf{Multi}
}
\label{tab:my-table}
\resizebox{\textwidth}{!}{%
\begin{tabular}{@{}c|ccccccccccccccc@{}}
\toprule
Setting                              & \multicolumn{15}{c}{Multi}                                                                                                                                                                                   \\ \midrule
Domain                               & \multicolumn{7}{c|}{Chemical Molecules}                               & \multicolumn{2}{c|}{Bioinformatics} & \multicolumn{3}{c|}{Social Networks}                   & \multicolumn{3}{c}{Computer Vision}   \\ \midrule
Datasets                             & MUTAG & BZR & COX2 & DHFR & PTC-MR & AIDS & \multicolumn{1}{c|}{NCI1} & ENZYMES  & \multicolumn{1}{c|}{DD}  & COLLAB & IMDB-BINARY & \multicolumn{1}{c|}{IMDB-MULTI} & Letter-low & Letter-high & Letter-med \\ \midrule
\# of clients                        & 1     & 1   & 1    & 1    & 1      & 1    & \multicolumn{1}{c|}{1}    & 1        & \multicolumn{1}{c|}{1}   & 1      & 1           & \multicolumn{1}{c|}{1}          & 1          & 1           & 1          \\
\# of graphs for training per client & 480   & 942 & 4000 & 800  & 1200   & 1800 & \multicolumn{1}{c|}{1800} & 480      & \multicolumn{1}{c|}{942} & 4000   & 800         & \multicolumn{1}{c|}{1200}       & 1800       & 1800        & 1800       \\ \bottomrule
\end{tabular}%
}
\end{table}

\subsubsection{Details of the Baseline Methods.}
We compare FedDense with five baselines. The details of these baselines are provided as follows.
\begin{itemize}
\item[$\bullet$] Local, where each client conducts model training locally without any communication with others.
\item[$\bullet$] FedAvg, a standard FGL approach that aggregates client models by averaging their local updates.
\item[$\bullet$] FedProx, where a regularization term in the loss function was proposed to handle system and statistical heterogeneity. The regularization term with importance weight $\mu$ is set to 0.01 in our experiments.
\item[$\bullet$] GCFL, which tackles non-IID graph data through a dynamic clustering technique based on GNN gradients to group clients with similar data distributions. Two hyper-parameters are determining the clustering results, i.e., $\epsilon_1$ and $\epsilon_2$. To guarantee the performance of GCFL, we use the same values in the original study where $\epsilon_1 = 0.05$ and $\epsilon_2 = 0.1$.
\item[$\bullet$] FedStar, a state-of-the-art FGL framework that decouples structural and feature learning and sharing across diverse graph domains. The structural embeddings for the structural channel are consistent with the original study. The concatenation of a degree-based embedding representing vertex degrees with one-hot encoding and a random walk-based positional embedding which is computed based on the random walk diffusion process, both with dimensions of 16.
\end{itemize}
\end{document}